# HINO: a BFO-aligned ontology representing human molecular interactions and pathways

Yongqun He*, Zoushuang Xiang

Department of Microbiology and Immunology, Unit for Laboratory Animal Medicine, and Center for Computational Medicine and Bioinformatics, University of Michigan Medical School, Ann Arbor, MI 48109, USA

**ABSTRACT**

Many database resources, such as Reactome, collect manually annotated reactions, interactions, and pathways from peer-reviewed publications. The interactors (*e.g.*, a protein), interactions, and pathways in these data resources are often represented as instances in using BioPAX, a standard pathway data exchange format. However, these interactions are better represented as classes (or universals) since they always occur given appropriate conditions. This study aims to represent various human interaction pathways and networks as classes via a formal ontology aligned with the Basic Formal Ontology (BFO). Towards this goal, the Human Interaction Network Ontology (HINO) was generated by extending the BFO-aligned Interaction Network Ontology (INO). All human pathways and associated processes and interactors listed in Reactome and represented in BioPAX were first converted to ontology classes by aligning them under INO. Related terms and associated relations and hierarchies from external ontologies (e.g., CHEBI and GO) were also retrieved and imported into HINO. HINO ontology terms were resolved in the linked ontology data server Ontobee. The RDF triples stored in the RDF triple store are queryable through a SPARQL program. Such an ontology system supports advanced pathway data integration and applications.

## INTRODUCTION

Various entities (e.g., nucleic acids, proteins, and small molecules) have participated in different biological interactions that can be grouped into pathways. Hundreds of databases of biological interactions and pathways exist on different websites. One of the most famous databases is Reactome that provides manually curated peer-reviewed pathway pathways and processes of human being and many other organisms [1]. As of November 13, 2013, Reactome contains 6,744 reactions and 1481 pathways for the human system (Ref: http://www.reactome.org/stats.html).

The pathway data in Reactome are stored in a relational database and can be downloaded in different formats, including flat file, BioPAX, SBML, and PSI-MITAB files. These formats are used primarily for the Reactome data exchange. For example, the BioPAX is an OWL-based biological pathway data exchange format [2]. The BioPAX ontology format represents basic high level pathways and associated interactions and entities participating in the interactions and pathways. Individual Reactome interactions and pathways are then represented as instances of BioPAX ontology classes. While the approach of listing individual pathways as instances in OWL is reasonable for data exchange, different levels of pathways (e.g., apoptosis pathways), reactions (e.g., phosphorylation), and entities (e.g., protein) are universals (or classes) instead of particulars (or instances). Given appropriate conditions, these pathways always occur. Therefore, they are more appropriately represented as classes instead of instances in the scope of ontology.

Formal ontologies are consensus-based controlled vocabularies of terms and relations with associated definitions, which are logically formulated and computer-understandable to promote automated reasoning. Hundreds of ontologies have been reported. To unify various ontologies, one approach to do is to align different domain ontologies under a top level ontology. The Basic Formal Ontology (BFO) is such a top level ontology [3]. BFO contains two branches, continuant and occurrent. The continuant branch represents time-independent entity such as material entity, and the occurrent branch represents time-related entity such as process. By aligning different domain ontologies under the two branches of BFO, we will be able to generate a comprehensive ontology-level structure that covers broad biological areas. Currently BFO has been used by over 100 domain ontologies, including those in biomedicine within the framework of the OBO Foundry [4].

The Interaction Network Ontology (INO) is a BFO-aligned ontology that targets on the representation of interactions and networks. INO was initially developed to classify and represent different levels of interaction keywords used for literature mining of genetic interaction networks [5]. In INO, a biological interaction is defined as a processual entity that has two or more participants (i.e., interactors) that have an effect upon one another. A biological network is a process that includes a network of at least two interactions. When an interaction network has a start(s) and an end(s), it is called a biological pathway.

In this study, we report our development of the Human Interaction Network Ontology (HINO) as an extension of INO. Currently, HINO targets for integrative representation of various human interactions and pathways recorded in Reacotme by using INO as the backbone, BFO as background, and multiple existing ontologies for ontology reuse and linkage. HINO is aimed to formally and logically represent various interactions from Reactome (and other resources in the future) as classes, supporting consistent usage of the interaction knowledge and automated reasoning. The integration of HINO with other ontologies allows better pathway data integration and analysis. HINO

To whom correspondence should be addressed: yongqunh@umich.edu

can also be used for making scientific queries and applied for advanced software development.

**METHODS**

**1.1 HINO development**

The format of HINO ontology is W3C standard Web Ontology Language (OWL2) [6]. For this study, many new HINO terms and logical definitions were manually added using the Protégé 4.2 OWL ontology editor [7]. Computational SPARQL queries were also developed to automatically generate many HINO terms or import existing ontology terms as detailed below.

**1.2 Transferring Reactome data to HINO**

Similar to HINO, the BioPAX also uses the OWL representation formalism. The OWL Description Logics (DL) enables the usage of reasoning to classify the ontological hierarchy following the axioms available. Therefore, we decided to use the BioPAX version of Reactome data that was downloaded from the Reactome website. A standard mapping strategy was designed to map the Reactome data in BioPAX to the INO format. Basically, different interaction entities, including protein, complex, small molecules, were mapped to under BFO term 'material entity'. Interaction and pathways were mapped to under BFO term 'process'. NCBI Taxonomy ontology (NCBI_Taxon) was used to generate organism taxonomies. CHEBI was used for small molecules. GO was used for cellular components and molecular functions. The Reactome BioPAX version provides detailed citation information (e.g., author name, article title, and journal) from PubMed peer-reviewed papers. In HINO, we have simplified the citation with only PubMed ID (i.e., PMID). The unique PMIDs can be used to easily retrieve detailed citation information from PubMed.

Based on the mapping strategy, a Java program based on OWL API (http://owlapi.sourceforge.net/) was generated to automatically transfer the Reactome data represented in BioPAX to the INO format. Finally, manual annotation and modification were used to adjust some transferred hierarchies to INO according to the INO design patterns.

**1.3 Generation of ontology term hierarchies in HINO using OntoFox**

Our strategy is to use existing ontologies and ontology terms to represent entities in Reactome whenever possible. This strategy led to the generation of many ontology terms in INO particularly from the NCBI Taxonomy ontology (NCBI_Taxon), Chemical Entities of Biological Interest (ChEBI) [8], Protein Ontology (PRO) [9], and the Gene Ontology (GO) [10]. Initially these terms were only represented as individual ontology IDs. To support better data representation and reasoning, we have used the OntoFox tool [11] to automatically retrieve the hierarchical structure of these ontology terms. In our importing, the full hierarchical structure of GO covering all designated GO IDs was imported. New ontology terms forming the hierarchies were then imported to HINO.

**1.4 Ontology storage, visualization, and analysis**

Hegroup Resource Description Framework (RDF) triple store (http://sparql.hegroup.org/) was used to store the RDF triple store. A RDF triple is a data entity composed of subject-predicate-object. A RDF triple store is a special database for the storage and retrieval of RDF triples. The Hegroup RDF triple store was generating using the Virtuoso RDF triple store management system. The Virtuoso server was used to read the HINO OWL file and automatically transfer the HINO OWL file to RDF triples. The details of HINO including HINO hierarchy and individual ontology data can be visualized using Ontobee, a linked ontology term server that uses the Hegroup RDF triple store [12]. A HINO SPARQL program was generated in the INO website (http://www.ino-ontology.org). SPARQL based methods were used for further HINO data query and analysis.

**1.5 Availability and license**

The INO and HINO are openly available in INO website: http://www.ino-ontology.org/. Both ontologies are available in Ontobee. The INO website URL on Ontobee is: http://www.ontobee.org/browser/index.php?o=INO. The Ontobee website for the HINO ontology is: http://www.ontobee.org/browser/index.php?o=HINO. Both ontologies have also been deposited to the NCBO BioPortal.

**RESULTS**

**2.1 Development of INO and HINO aligned with BFO**

The overall design of HINO is to generate an ontology of human interactions and pathways by aligning the ontology with the BFO structure (Fig. 1). Since INO is already aligned with BFO, HINO is aligned naturally with BFO as an INO extension ontology. INO fully imports BFO. In addition, INO and HINO import many terms from external ontologies, including the Relation Ontology (RO) [12], NCBI_Taxon, Ontology for Biomedical Investigations (OBI) [13], ChEBI, and Information Artifact Ontology (IAO; http://code.google.com/p/information-artifact-ontology/). The reason of aligning INO and HINO to BFO is that BFO is a well design top ontology that has been used by many ontologies including OBO Foundry ontologies such as GO and ChEBI. Our strategy allows INO/HINO seamlessly integrated with existing ontologies in the OBO Foundry library. Such alignment will allow the natural reuse of GO and ChEBI in HINO and facilitate the overall ontology integration. How these ontologies are aligned is indicated at the top level shown in Fig. 1. The following sections also indicate the power of the strategy of integrating existing ontologies.

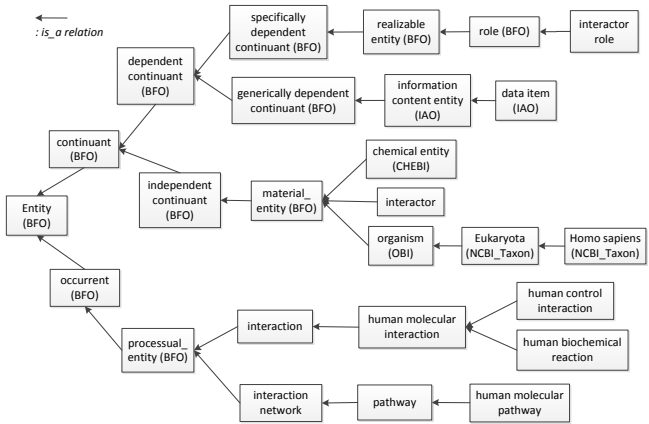

**Fig. 1.** General architecture of INO and HINO which align with BFO. Many terms are imported from existing ontologies as indicated by the ontology abbreviations cited in the parentheses. Those terms without indicated ontology are from either INO or HINO.

INO serves as a core that provides many superclasses for terms in HINO. Most terms in HINO come from the conversion of the Reactome human pathways from its BioPAX format. For proper structuring of HINO, many efforts were spent on designing specific patterns and implementing special mapping strategy between BioPAX and HINO.

As of November 13, 2013, HINO includes 38,435 classes, 57 object properties, 4 datatype properties, and 69 annotation properties. The latest only statistics can be obtained at the following Ontobee website: http://www.ontobee.org/ontostat.php?ontology=HINO.

### 2.2 Representation of species taxonomy in HINO

In Reactome BioPAX, the species taxonomy information is represented by BioPAX unique identifier, *e.g.*, the 'Homo sapiens' (human) has the unique identifier: http://www.reactome.org/biopax/48887#BioSource1. This type of identifiers does not support data integration. For example, the same class human may have different identifiers in different systems or data exchange formats. Such a design prevents appropriate data sharing and integration. The NCBI Taxonomy ontology (NCBI_Taxon) has been well recognized and used as an OBO Foundry library ontology. In HINO, we converted the BioPAX-specific representation of species to NCBI_Taxon representation.

Fig. 2 shows that in addition to human, the HINO or Reactome database also includes entities from the bacterium *M. tuberculosis* (Mtb) and virus Human immunodeficiency virus 1 (HIV1). Further analysis found that HINO includes the interaction data between human cells and *M. tuberculosis* or HIV1. For example, the HINO pathway term 'Response of Mtb to phagocytosis' represents how is a Mtb respond to the phagocytosis of a human phagocyte.

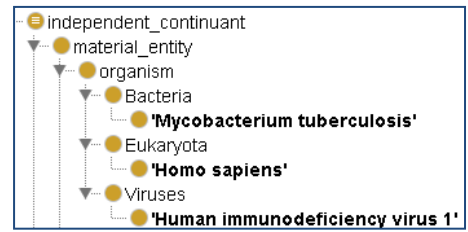

**Fig. 2.** The species and their taxonomic hierarchy used in the HINO. The NCBI_Taxon ontology was used to represent and classify the *Homo sapiens* (human, NCBITaxon_9606), *M. tuberculosis* (NCBITaxon_1773), and HIV1 (NCBITaxon_11676).

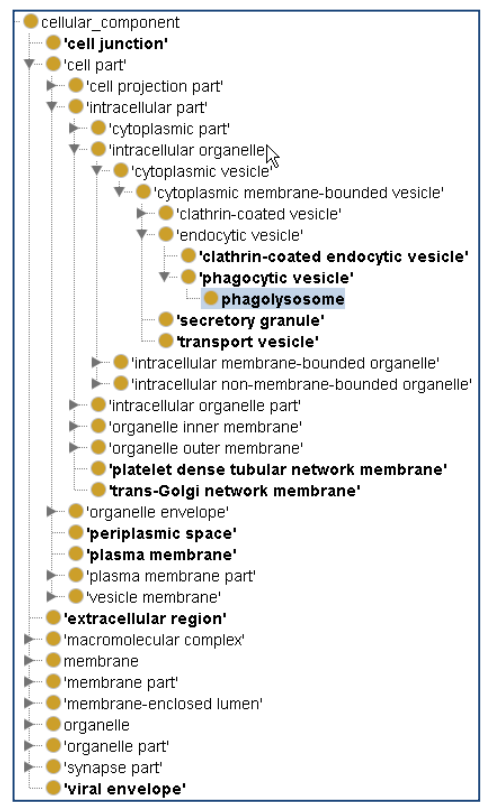

**Fig. 3.** The GO cellular component hierarchy used in the HINO.

### 2.3 Representation and analysis of cellular components and associated HINO entities

The Reactome database uses the GO:Cellular Components to reference the cellular locations of various interactions and pathways. The Reactome BioPAX uses its own IDs to represent the GO terms. In HINO, the GO ontology term URIs and annotations are directly imported from GO. In addition, the OntoFox tool was used to retrieve the original hierarchy of the two GO terms (Fig. 3). By doing so, we will be able to identify groups of cellular components (e.g., intracellular organelles) where various reactions and pathways are conducted. We can also easily identify the location where entities exist. For example, the GO term 'phagolysosome' is used by the term 'Exogenous Particulate antigen (Ag)' through the relation:

*located_in some phagolysosome*

where the term 'located_in' is a relation in the OBO Relation Ontology (RO).

### 2.4 CHEBI data analysis

Similar to the import of GO and GO ontology hierarchy to HINO, HINO also imports many CHEBI terms and their relational hierarchy for representation of small molecules used in HINO (Fig. 4). CHEBI is a community-based biological ontology focused on representation of chemical entities commonly used in the biological settings. The use of CHEBI term annotations and term hierarchy in HINO for representation of small molecules in human interactions and pathways in HINO allows clear visualization and analysis of which groups of chemical entities have been used in human interaction pathways.

For example, Fig. 4 shows that there are four divalent metal cations, including calcium (2+), copper (2+), magnesium (2+), and zinc (2+), which participate in various human molecular reactions and pathways collected in Reactome. Further exploration will lead to the identification of the reactions and pathways that use any or all of these metal cations.

### 2.5 HINO pathway representation as a set of molecular interaction processes

In HINO, a pathway is expressed a BFO:'processual entity' (Fig. 1). A pathway is composed of a set of molecular interactions, each of which is also a BFO:'processual entity' (Fig. 1). For example, Fig. 5 demonstrates how a complex pathway called 'Toll Like Receptor 4 (TLR4) Cascade (HINO_0022307) is generated in HINO. Human TLR4 is a Toll-like receptor protein that detects lipopolysaccharide (LPS) from Gram-negative bacteria. TLR4 is able to recruit four adapters to active two distinct signaling pathways [14]. It plays a critical role in the activation of the innate immune system after Gram-negative bacterial infection.

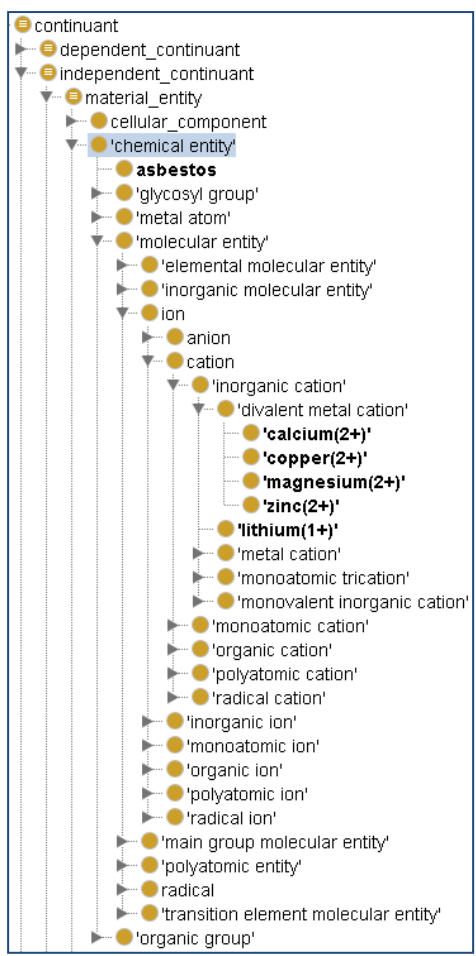

**Fig. 4.** The CHEBI hierarchy of small molecules used in HINO.

**Fig. 5.** A screenshot of an Ontobee display of a HINO term 'Toll Like Receptor 4 (TLR4) Cascade (HINO_0022307). The visualization includes the class label, term IRI, class hierarchy, superclasses & asserted axioms, as the uses of this term in the ontology HINO. The website also provides many web links to corresponding websites.

As shown in Fig. 5, the TLR4 cascase pathway in HINO has 9 separate "PathwaySteps" such as PathwayStep6776 as defined using the RO relation *has_part*. Each of the pathwaystep is either a subpathway (i.e., containing more than 1 interaction) or a single interaction. The order of these 9 pathway steps in the whole pathway is defined using the pathway order defined with the relation 'pathwayOrder'. The approach of using the "PathwayStep" comes from the method used in the BioPAX version of the human Reactome data (Fig. 5). Current approach is feasible. However, such a strategy may not be the best one. A better strategy is being designed and evaluated for better pathway information representation.

### 2.6 Ontobee visualization

Ontobee is a linked data server that is aimed to facilitate ontology visualization, query, and linkages [15]. This tool is primarily developed by our group. Ontobee has been used as the default linked data server for the majority of ontologies in the OBO Foundry library. In this study, we have also used Ontobee to display HINO ontology terms, their annotations, associated relations, and link them to associated ontologies. Fig. 5 provides an example of demonstrating the TLR4 cascase pathway. Although not shown in Fig. 5, the Ontobee HINO page also provides Reactome database ID and GO ID information for cross references, which allows a quick linkage to Reactome and GO for more specific graph representation and knowledge display.

In addition to ontology term visualization, Ontobee provides a RDF source code for any particular HTML web page for a specific ontology term. The return of a RDF source code to a web application request supports the Semantic Web data sharing [16]. Such a feature is not demonstrated in this paper but is illustrated in our previous report [15].

**Fig. 6.** A screenshot of a SPARQL query of subclasses of the INO term 'human molecular pathway' (INO_0000021). The query text is displayed on the top section and the query result is shown below.

### 2.7 SPARQL query results

The SPARQL Protocol and RDF Query Language (SPARQL) is an RDF query language. A specific web-based SPARQL program has been developed in the INO website for efficient and powerful query of the HINO information. For example, Fig. 6 shows that the execution of a simple SPARQL query was able to retrieve all subclasses of a HINO term "human molecular pathway' (INO_0000021)'. The query text is displayed on the top section of Fig. 6. Other SPARQL queries can also be developed for various queries and analyses.

## DISCUSSION

Hundreds of biological interaction and network pathways exist. The information among many of these databases is often redundant, and it is difficult to integrate the data. However, the integration of all these database knowledge is a big challenge. One major reason is that different databases use different formats to represent the pathway contents. While individual relational databases used to store pathway data are well designed in individual projects, it is difficult to share the data among different systems. Those standard data exchange formats including BioPAX and SBML are good for data exchange; however, these formats turn to store data in instances instead of class level representation. Our approach of representing pathway data using BFO-aligned ontology and incorporating as many reliable ontologies as possible provides a feasible way to eventually integrate different interaction pathways.

The main contribution of this report is that it demonstrates the feasibility and potential benefits of representing interactions and pathways as classes instead of instances. The current work reported in this manuscript uses Reactome to illustrate our principle idea that representing interactions and pathways as "classes-only" ontology aligning with BFO will make the knowledge in an interaction resourced more "interoperable". Currently, the knowledge in Reactome is not interoperable as it could be, and it is not directly connected with other ontologies either. By directly incorporating the Reactome interaction and pathway resources with ChEBI, NCBI_Taxon, and GO, we have made progress along the lines of ontological representation and interoperability.

Recently, Dr. He (Co-author of this paper) has initiated the development of a new Ontology of Genes and Genomes (OGG). The basis OGG information can be obtained at: http://www.ontobee.org/browser/index.php?o=OGG. The basic design was announced in recent email threads shown here: https://groups.google.com/forum/#!forum/ogg-discuss. Basically, Dr. He designed a novel framework that ontologically represents all individual genes of different organisms on one ontology system. The development of OGG is aimed to address a major bottleneck in biological ontology community, i.e. the lack of an ontology representing individual genes from human and other organisms. OGG integrates many resources of the USA National Center for Biotechnology Information (NCBI), including NCBI Entrez Genes, Genomes, and Taxonomy system. HINO includes the information of many human genes. The inclusion of OGG will likely further enhance the ontological representation of human interactions, pathways, and networks.

The HINO ontology representation approach can also be used to represent the interactions and pathways from other organisms in addition to human. The same strategy can be used to represent interaction and pathway resources in other resources, such as the Kyoto Encyclopedia of Genes and Genomes (KEGG) [17], BioCyc [18], NCI-Nature pathway database (http://pid.nci.nih.gov/), BioCarta [19]. These tasks will be implemented after current modeling of Reactome human interaction pathways optimized. When multiple data resources are used, a pitfall is that many redundant interactions and pathways will exist and should be avoided. We will also consider the incorporation of other ontologies in HINO, such as the Pathway Ontology [20].

One obvious application of HINO is its usage in interaction and pathway data integration. The integration is not only for integrating data in database resources like Reactome, but it is also for integrating knowledge in different ontologies, as demonstrated in this manuscript. If all human interaction and pathway data are uniquely represented in HINO, it would significantly avoid redundant work on interaction and pathway annotation and support data query and automated reasoning. The potential feature of avoiding representation redundancy from different resources by using the INO/HINO strategy is not inherently supported by BioPAX. BioPAX is designed to support data exchange of individual interaction/pathway data resources. The interactions and pathways in BioPAX are assigned with temporary instance identifiers, which cannot ontologically and inherently be used for class or universal-level stable representation and interoperable data reuse/comparison.

In addition to the data integration, the INO/HINO can also be used for other applications. For example, through integration with other ontologies such as GO, the INO/HINO can likely be used for development of new or improvement of existing statistical analysis methods for enhanced literature mining and gene expression data analysis.

## ACKNOWLEDGEMENTS

This project was supported by a NIH-NIAID grant (R01AI081062). This manuscript was originally submitted to the 4th International Conference on Biomedical Ontology (ICBO), Montreal, QC, Canada. We appreciate the two anonymous reviewers' constructive comments. We also appreciate the audiences who attended and commented to my poster presentation of this study in the conference (The short one-page summary of the presentation is available at: http://www2.unb.ca/csas/data/ws/icbo2013/papers/posters/icbo2013_submission_70.pdf). These comments have helped us to significantly improve this manuscript. We also look forward to more comments and collaborations on this project.